\documentclass{article}

\PassOptionsToPackage{numbers, compress}{natbib}

\usepackage[preprint]{neurips_2023}

\usepackage[utf8]{inputenc} 
\usepackage[T1]{fontenc}    
\usepackage{hyperref}       
\usepackage{url}            
\usepackage{booktabs}       
\usepackage{algorithm}
\usepackage{algpseudocode}
\usepackage[pdftex]{graphicx}
\usepackage{caption,tabularx,booktabs}
\usepackage{amsfonts}       
\usepackage{nicefrac}       
\usepackage{microtype}      
\usepackage{xcolor}         
\usepackage{amsmath}
\usepackage{graphicx}
\usepackage{amssymb}
\usepackage{soul}

\usepackage{amsmath}
\usepackage{amssymb}
\usepackage{bm}
\usepackage{enumitem}

\usepackage{amsmath}
\usepackage{amsfonts}
\usepackage{amssymb}
\usepackage{wrapfig}
\usepackage{subcaption}
\usepackage{multirow}
 \usepackage{mathtools} 

\usepackage{verbatim}

\usepackage{anyfontsize}

\usepackage{microtype}
\usepackage{graphicx}
\usepackage{booktabs} 
\usepackage{multirow}
\usepackage{amsmath,amssymb}
\usepackage{booktabs}
\usepackage{caption,subcaption}

\usepackage{xcolor}
\definecolor{mygreen}{HTML}{3cb44b}
\definecolor{skyblue}{HTML}{beffff}
\definecolor{lightgreen}{HTML}{90ee90}

\usepackage{tcolorbox}
\usepackage{enumitem}
\setitemize{itemsep=10pt,topsep=0pt,parsep=0pt,partopsep=0pt}
\usepackage{adjustbox}

\newcommand{\RN}[1]{%
	\textup{\lowercase\expandafter{\it \romannumeral#1}}%
}
\usepackage{tabu}

\usepackage{amsmath}

\newcommand{\beq}{\vspace{0mm}\begin{equation}}
\newcommand{\eeq}{\vspace{0mm}\end{equation}}
\newcommand{\beqs}{\vspace{0mm}\begin{eqnarray}}
\newcommand{\eeqs}{\vspace{0mm}\end{eqnarray}}
\newcommand{\barr}{\begin{array}}
\newcommand{\earr}{\end{array}}

\newcommand{\cv}[0]{{\boldsymbol{c}}}

\newcommand{\gv}[0]{{\boldsymbol{g}}}

\newcommand{\lv}[0]{{\boldsymbol{l}}}

\newcommand{\vv}{\boldsymbol{v}}

\newcommand{\xv}{\boldsymbol{x}}

\newcommand{\epsilonv}{\boldsymbol{\epsilon}}

\newcommand{\thetav}{\boldsymbol{\theta}}

\usepackage{color, colortbl}
\definecolor{Gray}{gray}{0.93}

\usepackage{lipsum}

\usepackage{color, colortbl}

\definecolor{emerald}{rgb}{0.31, 0.78, 0.37}

\newcommand{\MyColorBox}[2][red]%
{%
    \settowidth{\Width}{#2}%
    \colorbox{#1}%
    {%
        \raisebox{-\DepthReference}%
        {%
                \parbox[b][\HeightReference+\DepthReference][c]{\Width}{\centering#2}%
        }%
    }%
}

\definecolor{codegray}{gray}{0.9}

\title{Generate Anything Anywhere in Any Scene}

\author{
\textbf{{Yuheng Li, Haotian Liu, Yangming Wen, Yong Jae Lee}}\\\\
{University of Wisconsin--Madison}\\\
\url{https://yuheng-li.github.io/PACGen/}
}

\begin{document}

\maketitle

\begin{figure*}[h!]
\vspace{-20pt}
    \centering
    \includegraphics[width=\textwidth]{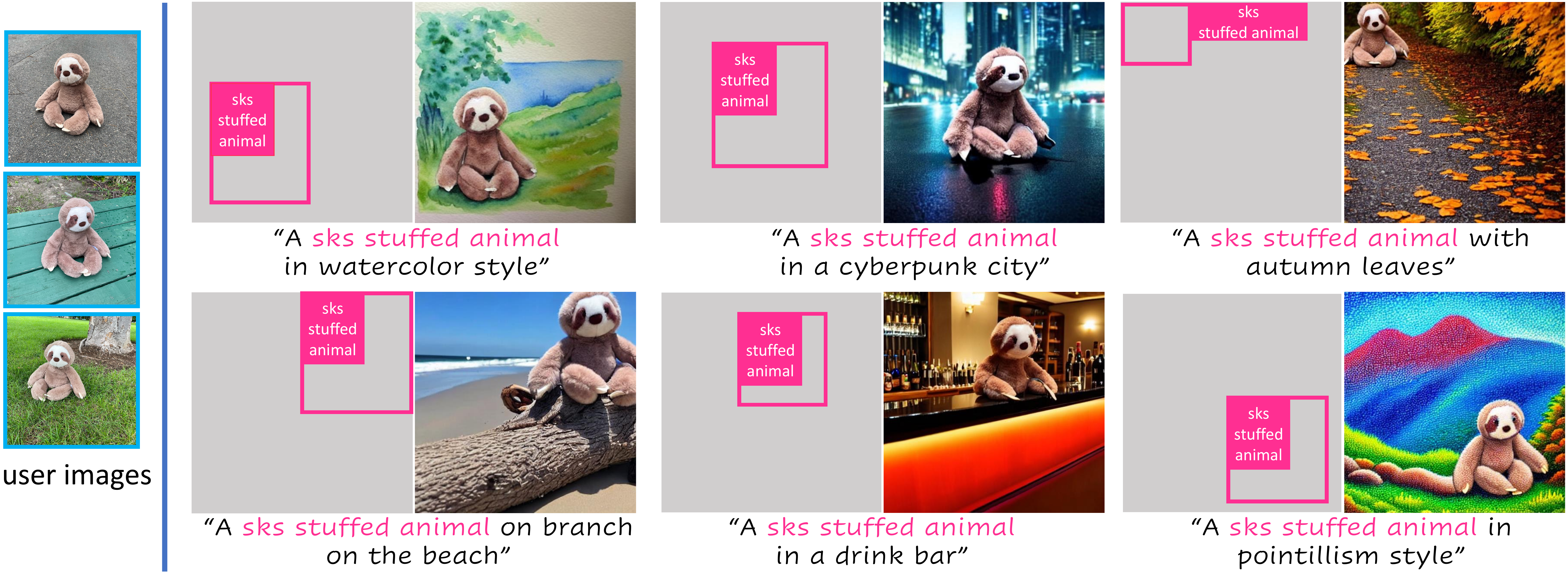}
    \caption{Given just a handful of user images (left), our model, PACGen, can generate the \emph{personalized} concept with both high fidelity and localization \emph{controllability} in novel contexts (right).}
    \label{fig:teaser}
\end{figure*}

\begin{abstract}
\vspace{-5pt}
Text-to-image diffusion models have attracted considerable interest due to their wide applicability across diverse fields. However, challenges persist in creating \emph{controllable} models for \emph{personalized} object generation. In this paper, we first identify the entanglement issues in existing personalized generative models, and then propose a straightforward and efficient data augmentation training strategy that guides the diffusion model to focus solely on object identity. By inserting the plug-and-play adapter layers from a pre-trained controllable diffusion model, our model obtains the ability to control the location and size of each generated personalized object.  During inference, we propose a regionally-guided sampling technique to maintain the quality and fidelity of the generated images. Our method achieves comparable or superior fidelity for personalized objects, yielding a robust, versatile, and controllable text-to-image diffusion model that is capable of generating realistic and personalized images. Our approach demonstrates significant potential for various applications, such as those in art, entertainment, and advertising design.
\end{abstract}

\vspace{-5pt}
\section{Introduction}
\vspace{-5pt}

The task of generating and manipulating photo-realistic images from textual descriptions has garnered significant attention in recent years. Solutions to this problem offer a wide range of potential applications in art, advertising, entertainment, virtual reality, and more. Recently, diffusion models \cite{DM,ddpm, ddim, GLIDE,DALLE2,LDM,Imagen}, together with cross-modal conditioning and large language models (e.g., CLIP \cite{CLIP} and T5-XXL \cite{T5}), have contributed significant advancements to this area.

While text-to-image diffusion models have significantly advanced both image realism and broadened the conceptual scope of what can be generated, there are still ongoing challenges. In many practical applications, \emph{controllable} models that can provide more fine-grained control beyond textual descriptions are highly sought after. For example, the freedom to generate \emph{any object} in \emph{any location} within \emph{any scene} is a desirable property. Such a model would enable a new degree of creative possibilities. For instance, advertisement designers could seamlessly position their products in any desired spot in an ad; likewise, an ordinary person could exhibit their most treasured toy in any location within the most prestigious showroom in Paris.

Researchers have been working diligently to create such controllable diffusion models. One approach entails \emph{personalization}, as seen in projects like Textual-Inversion~\cite{textualinv}, DreamBooth~\cite{ruiz2022dreambooth}, and Custom Diffusion~\cite{multiconcepy}. In these works, a pre-trained diffusion model is optimized or fine-tuned with a new personalized visual concept based on a handful of images provided by a user, to connect the visual concept to a unique word identifier. While these models can generate personalized items in any scene, they fall short in controlling the size and location of the generated objects. Another very recent research direction investigates \emph{localization-controllable} diffusion models~\cite{gligen,reco,controlnet,t2i,Composer,universal-bansal-arxiv2023}. In particular, GLIGEN~\cite{gligen} integrates a new adapter layer into each Transformer block of a pre-trained diffusion model to achieve size and location control. Although GLIGEN can generate objects according to a user-specified size and location information (e.g., provided by a bounding box or keypoints) and semantic category information provided via text, it cannot control the specific identity of the generated objects, rendering the model ineffective in dealing with personalized objects.

To tackle these challenges, in this paper, we propose a method, PACGen (Personalized and Controllable Text-to-Image Generation), that combines the key advantages of both research directions, leading to a more robust, versatile, and controllable text-to-image diffusion model capable of generating realistic and personalized images. We first show that personalization efforts like DreamBooth~\cite{ruiz2022dreambooth}, in the process of learning object identity, undesirably also entangle object identity with object location and size due to limited variability and quantity in the provided training data (e.g., the provided users' images are few and tend to be object-centric). This fundamental issue presents difficulties for location and size control in personalized generative models.

To address this issue, we propose a straightforward data augmentation technique. Specifically, when finetuning the pretrained diffusion model for personalization, we randomly resize and reposition the user provided training images in order to disentangle object identity from its size and location factors. During inference, we can simply plug in a pre-trained GLIGEN's adapter layers to control the object's size and location. This strategy enables  PACGen to associate the word identifier solely with the object's identity and allows it to control its size and location without any further finetuning. However, we find that some undesirable artifacts (like collaging effects) can be introduced from data augmentation. We thus also introduce a regionally-guided sampling technique to ensure the overall quality and fidelity of the generated images.

In sum, our main contributions are: 1) we identify and propose a solution to the entanglement issues in existing personalization efforts; 2) a novel text-to-image diffusion model that provides both personalization and controllability; and 3) our quantitative and qualitative results demonstrate that PACGen matches or exceeds the high fidelity of existing personalized models, while also providing localization controllability which is lacking in existing personalized models.

\vspace{-5pt}
\section{Related Work}
\vspace{-5pt}



\textbf{Text-to-Image Generation Models.} Text-to-image generation has been an active research area for years, with initial focus on GAN~\cite{GANs2016}-based techniques~\cite{stackgan,stackgan++,xu2018attngan,zhu2019dm}. However, these methods faced challenges when applied to general-domain datasets or scenes, in part to the training instabilities of GANs. Subsequent approaches like VQGAN~\cite{VQGAN}, DALL-E~\cite{ramesh2021zero}, and CogView~\cite{ding2021cogview} integrated Transformers~\cite{vaswani2017attention} to produce images with greater fidelity. The recent emergence of diffusion models~\cite{DM,ddpm,ddim,LDM}, capitalized on the power of cross-modal language-image representations such as CLIP~\cite{CLIP} and large language models like T5-XXL~\cite{T5}, has facilitated significant improvements in text-driven image generation and manipulation. Notable examples including GLIDE~\cite{GLIDE}, DALL-E2~\cite{DALLE2}, Imagen~\cite{Imagen}, and Stable Diffusion~\cite{Imagen}. Our work advances this line of work by providing both personalization and localization controllability to text-to-image generation models.


\textbf{Conditional Text-to-Image Diffusion Models.} Observing that text-to-image methods lack precise control over the generated content, recent work has explored a variety of additional conditioning factors beyond text in text-to-image diffusion models. For example, GLIGEN~\cite{gligen}, ControlNet~\cite{controlnet}, T2I-Adapter~\cite{t2i}, and Composer~\cite{Composer} can condition the diffusion model on text as well as bounding boxes, poses, depth maps, edge maps, normal maps, and semantic masks. These works demonstrate a simple and effective approach to injecting the additional controlling information via a trainable plug-and-play module without affecting the pre-trained diffusion model's original capabilities. However, these methods lack the ability to control the subject identity for personalized image generation. 

\textbf{Personalized and Controllable Image Generation.} A key objective in image generation is to enable personalized and controllable manipulation. GAN Inversion \cite{ganinv} accomplishes this by converting a given image into a corresponding latent representation of a pretrained GAN \cite{styleGAN,bigGAN}. In the realm of diffusion-based text-to-image models, DALL-E2 \cite{DALLE2} has shown promise by mapping images into CLIP-based codes, though it struggles to capture unique details of personalized objects. Textual Inversion \cite{textualinv} inverts a user-provided image into a word vector for personalized image generation, demonstrating impressive flexibility of the textual embedding space. However, it falls short when tasked with rendering the same subject in new contexts \cite{ruiz2022dreambooth}. In contrast to optimizing a word vector, DreamBooth  \cite{ruiz2022dreambooth} embeds and finetunes the given subject instance into the entire text-to-image diffusion model. Further, Custom Diffusion \cite{multiconcepy} composes multiple new concepts by optimizing only a small subset of weights in the diffusion model's cross-attention layers. Despite these advances, maintaining semantic coherence while exerting fine-grained control over the synthesized images still remains a complex challenge. Our work tackles this by introducing fine-grained localization control from GLIGEN \cite{gligen} into the customized fine-tuning process of DreamBooth \cite{ruiz2022dreambooth} to enable both personalized and controllable capabilities in text-to-image generation.

\vspace{-5pt}
\section{Background on DreamBooth and GLIGEN}
\vspace{-5pt}

In this section, we briefly introduce two key components of our model, DreamBooth~\cite{ruiz2022dreambooth} and GLIGEN~\cite{gligen}, and discuss their limitations, to motivate our work on combining their strengths for a more powerful text-to-image diffusion model that provides both \emph{personalization} and \emph{controllability}.

\vspace{-1pt}
\subsection{DreamBooth}
\vspace{-1pt}

Personalization of text-to-image diffusion models has emerged as a hot topic in recent research. DreamBooth~\cite{ruiz2022dreambooth} is a notable contribution. It works by establishing a link between a personalized visual concept with a few (3-5) user images and an identifier in the text vocabulary. To achieve this, DreamBooth employs a combination of a rare token $V*$ (e.g., \texttt{sks}) together with the coarse category name (e.g., ``toy''), to represent the personalized concept. To incorporate the novel concept into a pretrained diffusion model, the model is finetuned with a prompt containing the concept (e.g., \texttt{a photo of a [sks] [class name]}) using the regular denoising diffusion objective:
\begin{align}\label{eq:ldm_loss}
\min_{\thetav} \mathcal{L} = \mathbb{E}_{\xv, \epsilonv \sim  \mathcal{N}(\mathbf{0}, \mathbf{I}), t} \big[ \|  \epsilonv -   f_{\thetav}(\xv_t, t, \cv) \|^2_2 \big],
\end{align}
where $t$ is sampled from time steps $\{1, \cdots, T\}$, $\xv_t$ is the step-$t$ noisy variant of image $\xv$, and $ f_{\theta} (*, *, \cv)$ is the denoising network conditioned on text feature $\cv$.   

To mitigate language drift~\cite{lee2019countering,lu2020countering} and maintain output diversity, DreamBooth is trained to reconstruct other (generated) images of the same class as the personalized item for half of the training time. After adapting the pretrained model's weights, DreamBooth can combine the personalized concept with the identifier $V*$ to generate new images, e.g., placing the personalized concept into new contexts.

\vspace{-1pt}
\subsection{GLIGEN}
\vspace{-1pt}

Another line of topical research is investigating controllable generation in pretrained diffusion models. In particular, GLIGEN~\cite{gligen} enhances the functionality of text-to-image diffusion models by providing additional control through bounding boxes, keypoints, and other conditioning features. It does this by freezing all of the original weights of a pretrained diffusion model and inserting a new trainable adaptor, a self-attention layer that takes in the additional conditions, into each Transformer block: 
\begin{align}
& \vv = \vv + \text{SelfAttn}(\vv)	 
\label{eq:ldm_sa}\\
& \vv = \vv + \tanh(\gamma) \cdot \text{TS}(\text{SelfAttn}([\vv, \gv]))
\label{eq:gated-self-attention}\\
& 
\vv = \vv + \text{CrossAttn}(\vv, \cv) 
\label{eq:ldm_ca}
\end{align}
Eqs~\ref{eq:ldm_sa} and \ref{eq:ldm_ca} represent the original diffusion model layers, where $\vv$ and $\cv$ denote the visual and text prompt tokens, respectively. Eq.~\ref{eq:gated-self-attention} is the new trainable layer introduced by GLIGEN, with $\gv$ representing the grounding tokens that contain additional controlling information, such as bounding boxes. The token selection operation, $\text{TS}(\cdot)$, considers only visual tokens, and $\gamma$ is a learnable scalar that acts like a gate (i.e., how much of the original pretrained model's capabilities should be retained).

After training with a relatively small amount of new data, GLIGEN (with box conditioning) can determine the location and size of noun entities in the prompt. Specifically, the prompt feature $\cv$ (e.g., \texttt{a photo of a cat and dog}) is fed into the cross-attention layer in Eq.~\ref{eq:ldm_ca}. The grounding tokens $\gv$ (e.g., one for \texttt{dog} and one for \texttt{cat}) are used in Eq.~\ref{eq:gated-self-attention} to control each object's location. At a high-level, each grounding token is derived from:
\begin{align}
\gv = f(e,\lv), 
\label{eq:grounding_token}
\end{align}
where $e$ is the feature of the grounded noun text (e.g., \texttt{cat}) using the same text encoder as $\cv$, $\lv$ is its bounding box, and $f(\cdot)$ is a simple MLP. Importantly, GLIGEN exhibits open-set  generation capabilities, even when finetuned on a limited number of categories (e.g., COCO~\cite{coco}). For example, it can control the location and size of entities like \texttt{hello kitty} even though it is not a category in the COCO dataset. It does this by learning a feature correspondence between $\cv$ and $\gv$ so that it can use the grounding condition in a category-agnostic way.  More details can be found in~\cite{gligen}.

\vspace{-5pt}
\section{Approach}
\vspace{-5pt}

While both DreamBooth and GLIGEN have brought significant advances to text-to-image diffusion models, they  possess certain limitations. Specifically, DreamBooth is capable of generating images that incorporate personalized concepts, but lacks the ability to control their precise location and size. On the other hand, GLIGEN offers greater control over object size and location, but cannot handle personalized objects. These limitations call for an approach which combines their strengths, enabling \emph{personalized} object generation with \emph{controlled} placement.

\begin{figure*}[t!]
    \centering
    \includegraphics[width=0.99\textwidth]{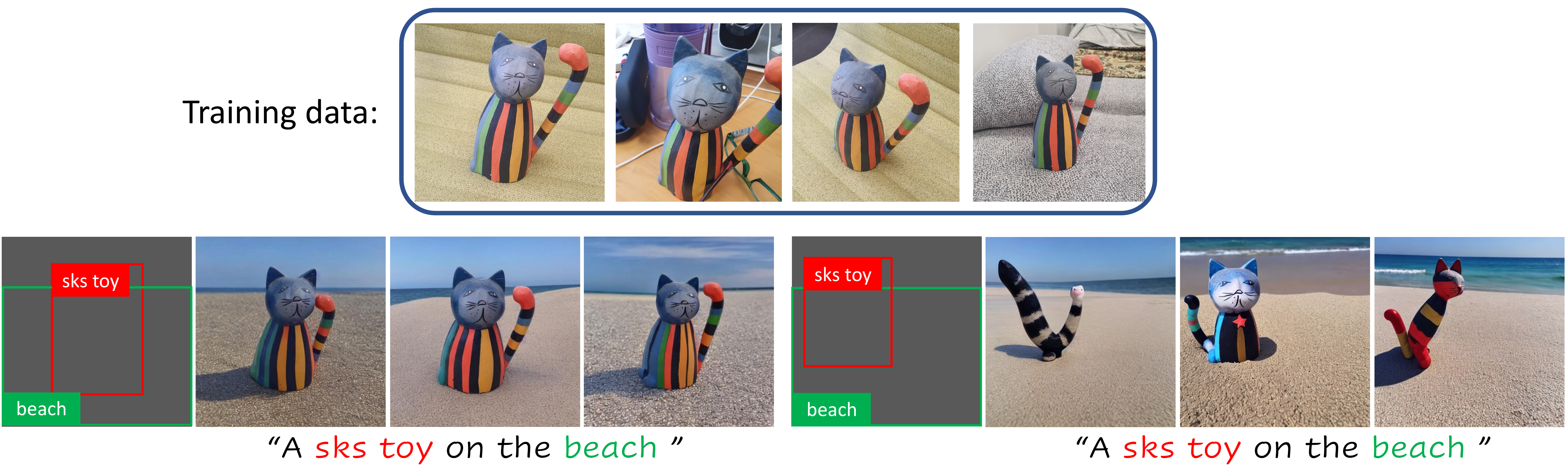}
    \caption{A naive combination of DreamBooth and GLIGEN.  (Left) The model generates accurate identities when the bounding box size and location roughly match those of the training distribution. (Right) However, it fails when the size and location fall outside the training distribution.}
    \label{fig:problem}
\end{figure*}

In the ensuing, we first identify an issue with DreamBooth, namely that it entangles object identity with other factors like location and size.  We then propose a simple remedy using data augmentation.  Finally, to improve image quality and fidelity, we propose a regionally-guided sampling technique.

\vspace{-1pt}
\subsection{DreamBooth Incorrectly Entangles Object Identity and Spatial Information}
\vspace{-1pt}


To enable spatial control over personalized concepts, a straightforward but naive solution would be to plug in GLIGEN's controllable layers into a pretrained DreamBooth model (this is possible when both GLIGEN and DreamBooth build upon the same diffusion model architecture, e.g., Stable Diffusion). For example, suppose DreamBooth is trained with the toy cat images in Fig.~\ref{fig:problem} with its learned unique identifier $V* =\texttt{sks}$. During inference, the corresponding grounding token $\gv$ can be computed by combining the text feature of \texttt{sks toy cat} and its location $\lv$ into Eq.~\ref{eq:grounding_token} to control its location using the pre-trained GLIGEN layers.


\begin{figure*}[t!]
    \centering
\includegraphics[width=0.99\textwidth]{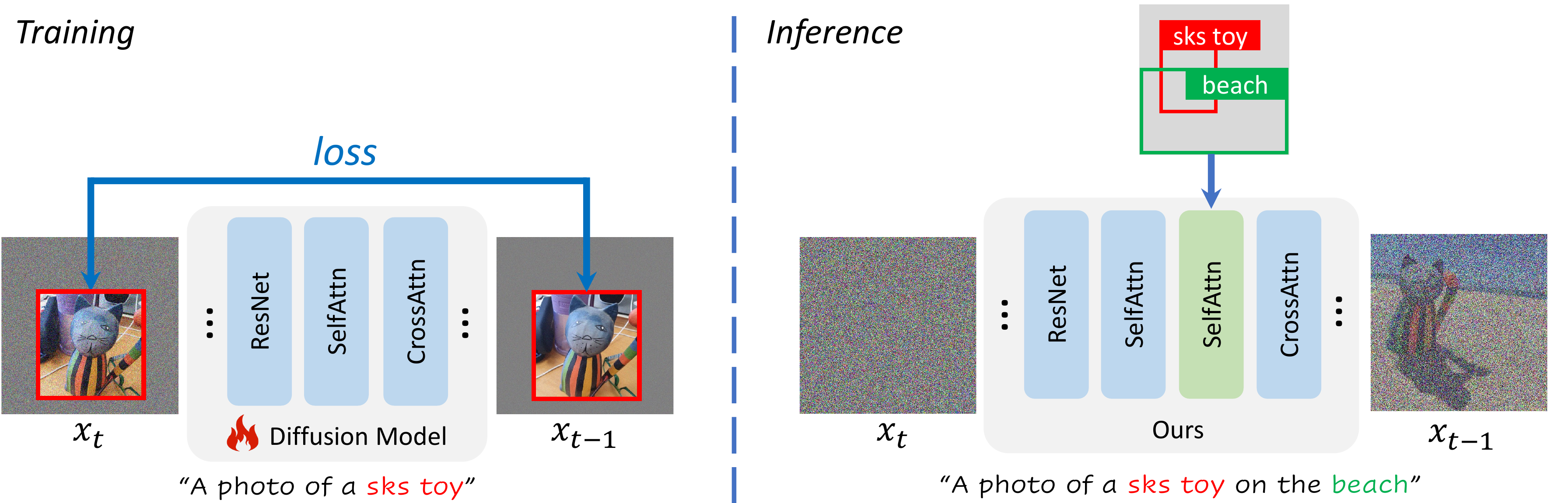}
    \caption{
By incorporating a data augmentation technique that involves aggressive random resizing and repositioning of training images, PACGen effectively disentangles object identity and spatial information in personalized image generation. 
}
    \label{fig:approach_training}
\end{figure*}

Fig.~\ref{fig:problem} shows that when the specified bounding box size and location roughly follow the size and location of the toy cat in the four user-provided training images, the generated images correctly contain the object identity (left). However, if the bounding box size and location fall outside the training distribution, the generated object's identity can deviate significantly from the original object (right). This suggests that DreamBooth not only learns the object's identity but \emph{also its location and size} within the training data; i.e., it overfits to the limited training data and incorrectly learns to \emph{entangle identity and spatial information}.

\begin{figure*}[h!]
    \centering
    \includegraphics[width=0.95\textwidth]{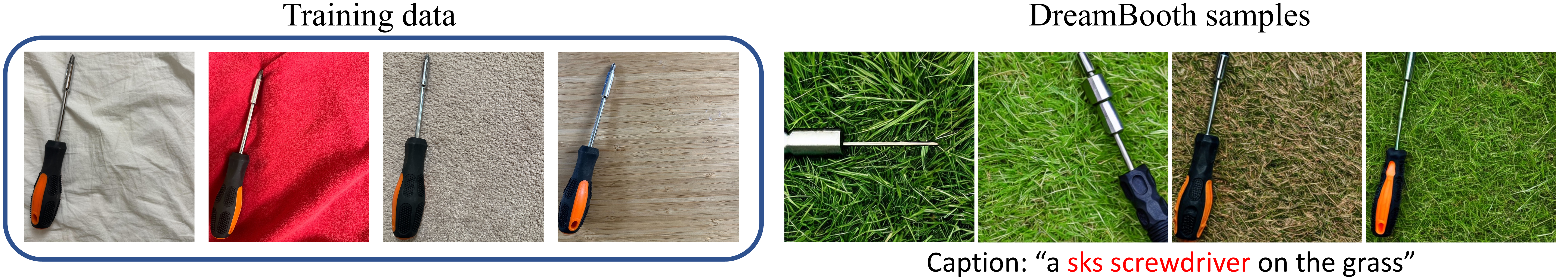}
    \caption{DreamBooth incorrectly learns to entangle object identity with spatial information during training. It generates the correct identity only when the location matches the training distribution.}
    \label{fig:problem2}
\end{figure*}

To further substantiate this hypothesis, we create a toy dataset comprising four screwdriver images, all positioned on the left side of the image, as depicted in Fig.~\ref{fig:problem2}. Upon training DreamBooth on this dataset, we discover that the model can accurately generate the object identity only when the generated objects are situated in the same location as the training data. In other locations, DreamBooth generates a screwdriver that is different from that in the training data.

\vspace{-1pt}
\subsection{Disentangling Identity and Spatial Information in Personalized Image Generation}
\vspace{-1pt}

A possible remedy to disentangle object identity and spatial information would be to ask the user to provide many more photos of their personalized concept in various locations and sizes. However, in many cases, this would be highly impractical and undesirable.  We thus present a simple training strategy which enables disentangled control over object identity and spatial information, even when there are only a handful of training images. Rather than directly use the handful of user-provided training images as-is to train the diffusion model, we apply aggressive data augmentation to transform the size and location of the personalized concept.  More formally, denote a training image as $x$ and the transformed input to the diffusion model as $y$. We obtain $y$ using the following transformation:
\begin{align}\label{eq:aug}
y = \text{aug}(s, p, g(x)),
\end{align}
where $g(\cdot)$ represents the combined rescale and center crop operation, ensuring that the user image conforms to the expected input size of the pretrained diffusion model (e.g., $512 \times 512$) -- this is needed since a user image can be of any arbitrary resolution.  $\text{aug}(s, p, \cdot)$ is a random resize and position operation, which randomly resizes (with resizing factor randomly sampled from $[s,1]$, $0<s<1$) and inserts the user provided image $g(x)$ into a clear gray image at sampled location $p$. The training objective is still the denoising loss (Eq.~\ref{eq:ldm_loss}) but only applied to the relevant image region.  Fig.~\ref{fig:approach_training} (left) illustrates this training methodology, where the loss is applied only to the pixels within the red box. 




\subsection{Improving Image Quality and Fidelity during Inference}

\begin{figure*}[t!]
    \centering    \includegraphics[width=0.95\textwidth]{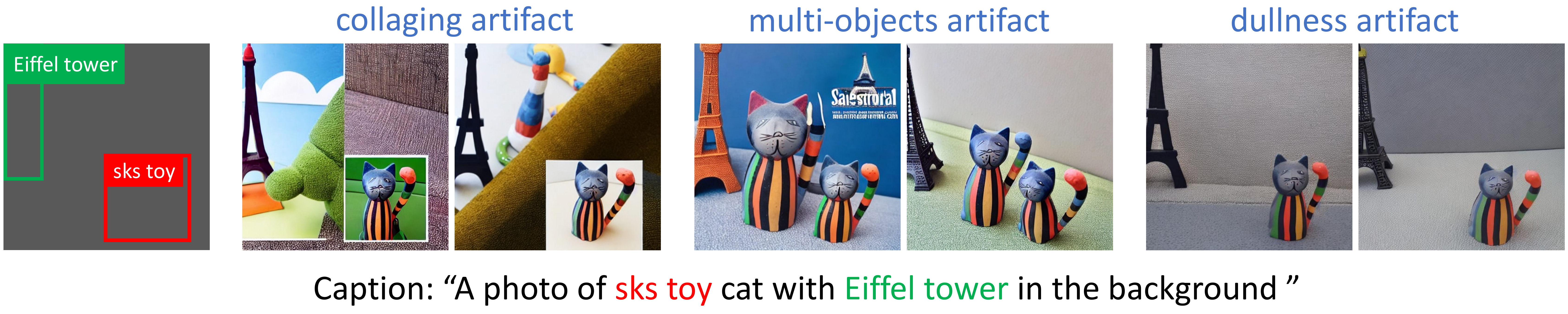}
    \caption{Data augmentation sometimes introduces collaging, multi-object, and dullness artifacts.    
    }
    \label{fig:artifact}
\end{figure*}

With the above data augmentation strategy, PACGen, which integrates DreamBooth with GLIGEN's controllable layers, can now effectively generate objects with accurate identity and position control. However, as illustrated in Fig.~\ref{fig:artifact}, we sometimes observe three types of artifacts introduced by the augmentation technique: collaging (examples 1, 2), multiple objects where the extra object(s) appears in unspecified locations (examples 3 to 4), and a gray dullness effect in which the generated image has an overall dull and plain gray color (examples 5 and 6). We hypothesize that these model behaviors stem from the fact that the loss function is applied exclusively to the image region (pixels within the red box in Fig.~\ref{fig:approach_training}), resulting in an undefined loss for areas outside this region. The collaging and dullness effects may further be attributed to the model's ability to `see' a sharp boundary and uniform value in the external region during training due to the receptive field of convolution and self-attention.

We introduce a regionally guided sampling technique to address these issues, without any further training, inspired by negative prompting. Specifically, in standard text-to-image diffusion sampling, we can denote the input text as a positive prompt $p_{\mathrm{postive}}$. Classifier-free guidance~\cite{cfg} was introduced to improve image fidelity and quality by utilizing a null text. More recently, the research community has discovered that replacing the null text with a negative prompt $p_{\mathrm{negative}}$ (i.e., undesirable properties in the output, such as \texttt{bad quality}, \texttt{low resolution}) can be more effective:  
\begin{align}\label{eq:prompt}
\mathbf{\epsilon}_{\mathrm{aggregate}} = \mathbf{\epsilon}_{\mathrm{negative}} + s * (\mathbf{\epsilon}_{\mathrm{positive}} - \mathbf{\epsilon}_{\mathrm{negative}})
\end{align}
where $\mathbf{\epsilon}_{\mathrm{positive}}$ and $\mathbf{\epsilon}_{\mathrm{negative}}$ are noise predicted by the corresponding prompts, and $s$ is a scalar guidance scale (typically $s>1$). The final noise $\mathbf{\epsilon}_{\mathrm{aggregate}}$ is used in the denoising sample step.  



We introduce two additional prompts: a multi-object suppression prompt $p_{\mathrm{suppress}}$ and a diversity encouraging prompt $p_{\mathrm{diverse}}$. We define $p_{\mathrm{suppress}}$ as the combination of the identifier and class name (e.g., \texttt{sks toy}) used during training. It aims to suppress the learned concept outside the user-provided layout (bounding box) region. For $p_{\mathrm{diverse}}$, we choose \texttt{a high quality, colorful image} as the default prompt. Since we do not want to suppress the object inside the layout region, and we also find that applying the diversity prompt $p_{\mathrm{diverse}}$ outside the layout region is sufficient to encourage the whole image to be high quality and colorful, we define our regional negative prompt as:
\begin{align}\label{eq:negprompt}
\mathbf{\epsilon}_{\mathrm{negative}} = \mathbf{m} * \mathbf{\epsilon}_{\mathrm{negative}} + (\mathbf{1} - \mathbf{m}) * (\mathbf{\epsilon}_{\mathrm{negative}} + \mathbf{\epsilon}_{\mathrm{suppress}} - \mathbf{\epsilon}_{\mathrm{diverse}})
\end{align}
where $\mathbf{m}$ is a mask whose value is 1 inside the $V*$ object region, which can be derived from the input layout; $\mathbf{1}$ is an all one mask. Note that we do not introduce any additional prompt engineering work for the user, since $p_{\mathrm{diverse}}$ is fixed for all personalized concepts and  $p_{\mathrm{suppress}}$ can be computed automatically based on the user-provided layout information. Alg.~\ref{algo:our_algorithm} shows the inference procedure.

\begin{algorithm}
    \footnotesize
    \caption{Inference Algorithm}
    \label{algo:our_algorithm}    
    \begin{algorithmic}[1]
    \State \textbf{Input}: Personal-GLIGEN $\phi_{\theta}$, layout $l$, sampler $\mathcal{S}(\mathbf{x}_t, \mathbf{\epsilon})$, guidance scale $s$
    \State \quad \quad \quad Prompts: $p_{\mathrm{positive}}$, $p_{\mathrm{negative}}$,  $p_{\mathrm{suppress}}$, $p_{\mathrm{diverse}}$

    \State \textbf{Sample} a noise $\mathbf{x}_T$
    \State \textbf{Extract} regional mask $\mathbf{m}$ from $l$
    \For{$t=T,\cdots,1$}
        \State $\mathbf{\epsilon}_{\mathrm{positive}} = \phi_{\theta}(\mathbf{x}_t, p_{\mathrm{positive}}, t, l)$;~~~ $\mathbf{\epsilon}_{\mathrm{negative}} = \phi_{\theta}(\mathbf{x}_t, p_{\mathrm{negative}}, t, \mathrm{NULL})$
        \State $\mathbf{\epsilon}_{\mathrm{suppress}} = \phi_{\theta}(\mathbf{x}_t, p_{\mathrm{suppress}}, t, \mathrm{NULL})$;~~~
        $\mathbf{\epsilon}_{\mathrm{diverse}} = \phi_{\theta}(\mathbf{x}_t, p_{\mathrm{diverse}}, t, \mathrm{NULL})$
        \State $\mathbf{\epsilon}_{\mathrm{negative}} = \mathbf{m} * \mathbf{\epsilon}_{\mathrm{negative}} + (\mathbf{1} - \mathbf{m}) * (\mathbf{\epsilon}_{\mathrm{negative}} + \mathbf{\epsilon}_{\mathrm{suppress}} - \mathbf{\epsilon}_{\mathrm{diverse}})$
        \State $\mathbf{\epsilon}_{\mathrm{aggregate}} = \mathbf{\epsilon}_{\mathrm{negative}} + s * (\mathbf{\epsilon}_{\mathrm{positive}} - \mathbf{\epsilon}_{\mathrm{negative}})$
        \State $\mathbf{x}_{t-1} = \mathcal{S}(\mathbf{x}_t, \mathbf{\epsilon}_{\mathrm{aggregate}})$
    \EndFor
    \State \textbf{Output}: $\mathbf{x}_0$
    \end{algorithmic}
\end{algorithm}

\vspace{-5pt}
\section{Experiments}
\vspace{-5pt}

In this section, we evaluate PACGen's personalized controllable text-to-image generation capabilities both quantitatively and qualitatively, and ablate our design choices.

\textbf{Datasets.} We use the same dataset as DreamBooth~\cite{ruiz2022dreambooth}, which contains 30 subjects, toy-cat example from~\cite{textualinv}, and screwdriver example we created in Fig.~\ref{fig:problem2}. Together we have 32 objects covering diverse common categories such as backpacks, stuffed animals, dogs, cats, sunglasses, cartoons, etc.

\textbf{Baselines.} We compare to four baselines: (1)  DreamBooth~\cite{ruiz2022dreambooth} and (2) Textual Inversion~\cite{textualinv} are pioneering approaches in personalized item generation with optimization-based techniques; (3) Custom Diffusion~\cite{multiconcepy}, a computationally efficient technique that optimizes only a small subset of cross-attention weights; and (4) GLIGEN~\cite{gligen}, which enables additional controllability in text-to-image diffusion models. It has multiple modalities and we use GLIGEN (image) which embeds a reference image using an image encoder to control its location using bounding box information. The original DreamBooth code is unavailable and has been studied using Imagen~\cite{Imagen}. Therefore, we utilize the code from diffusers~\cite{diffusers} and Stable Diffusion~\cite{LDM} as the pre-trained model. For the other three baselines, we use their official implementations, all of which are built upon Stable Diffusion.

\textbf{Metrics.} Following~\cite{multiconcepy}, we use three evaluation criteria: (1) Image alignment, which refers to the visual resemblance between the generated image and target concept, measured using similarity in CLIP image feature space~\cite{CLIP}; (2) Text alignment, which measures the correspondence between the generated image and given prompt (we use prompts from the DreamBooth dataset), computed using text-image similarity in CLIP feature space; and (3) KID~\cite{kid}, which is used to evaluate the forgetting of existing related concepts when compared to the distribution of real images retrieved from LAION-400M~\cite{laion}. In addition, we also compute (4) Object alignment, similar to image alignment, but after cropping out the object region using an object detector~\cite{SAM} and (5) IOU, which measures the spatial overlap between the generated object's location and the provided bounding box. 

\vspace{-2pt}
\subsection{Single Object Results}
\vspace{-2pt}

We first evaluate our method's ability to generate single object images.  Following~\cite{multiconcepy}, we use 20 prompts and generate 50 images per prompt.  To create the bounding box inputs that our method needs for generation in an automatic and fair manner, we first generate results using DreamBooth and then fire an object detector~\cite{SAM}. Since all baselines except GLIGEN cannot take in a bounding box as input, we only report IOU for GLIGEN and DreamBooth (the latter is simply for a lower-bound performance reference). Since GLIGEN takes in image CLIP features from one inference image, we average the features of all (3$\sim$5) reference images and provide it the resulting average feature.

\begin{table}[!t]
\centering
\setlength{\tabcolsep}{5pt}
\small
\begin{tabular}{ll c c c c c}
\toprule
& & \multicolumn{1}{c}{\textbf{Text}} & \multicolumn{1}{c}{\textbf{Image}} & \multicolumn{1}{c}{\textbf{Object}} & & \\
& \textbf{Method} & \multicolumn{1}{c}{\textbf{alignment} $(\uparrow)$} & \multicolumn{1}{c}{\textbf{alignment} $(\uparrow)$} & \multicolumn{1}{c}{\textbf{alignment} $(\uparrow)$} & \textbf{KID} $(\downarrow)$ & \textbf{IOU} $(\uparrow)$ \\
\midrule
\multirow{5}{*}{\shortstack[c]{\textbf{Single-}\\\textbf{Concept}}} & DreamBooth~\cite{ruiz2022dreambooth} & 0.779 & 0.740 & 0.825 & 22.580 & 0.395 \\
& Textual Inversion~\cite{textualinv} & 0.643 & 0.741 & 0.766 & \textbf{17.037} & - \\ 
& Custom Diffusion~\cite{multiconcepy} & 0.785 & \textbf{0.776} & 0.818 & 19.301 & - \\
& GLIGEN (image)~\cite{gligen} & 0.691 & 0.775 & 0.799 & - & 0.787 \\
& PACGen (Ours) & \textbf{0.794} & \textbf{0.776} & \textbf{0.828} & 19.582 & \textbf{0.795} \\
\midrule
\multirow{2}{*}{\shortstack[c]{\textbf{Multi-}\\\textbf{Concept}}} & Custom Diffusion~\cite{multiconcepy} & \textbf{0.841} & 0.690 & 0.732 & - & - \\
& PACGen (Ours) & 0.818 & \textbf{0.724} & \textbf{0.744} & - & \textbf{0.642} \\
\bottomrule
\end{tabular}
\caption{(Top rows) Single-concept evaluation averaged across 32 datasets. Our method consistently achieves the best semantic and identity alignment, while also enabling position control. Our method achieves similar KID score as Custom Diffusion, and is better than DreamBooth. Textual Inversion has the lowest KID as it does not update the model.
(Bottom rows) Multi-concept evaluation averaged across the 3 composition pairs. Our method performs worse/better/better than Custom Diffusion~\cite{multiconcepy} for text/image/object alignment. Importantly, ours provides position control, which~\cite{multiconcepy} lacks.
}
\label{table:main}
\vspace{-10pt}
\end{table}

\begin{figure*}[t!]
    \centering
    \includegraphics[width=\textwidth]{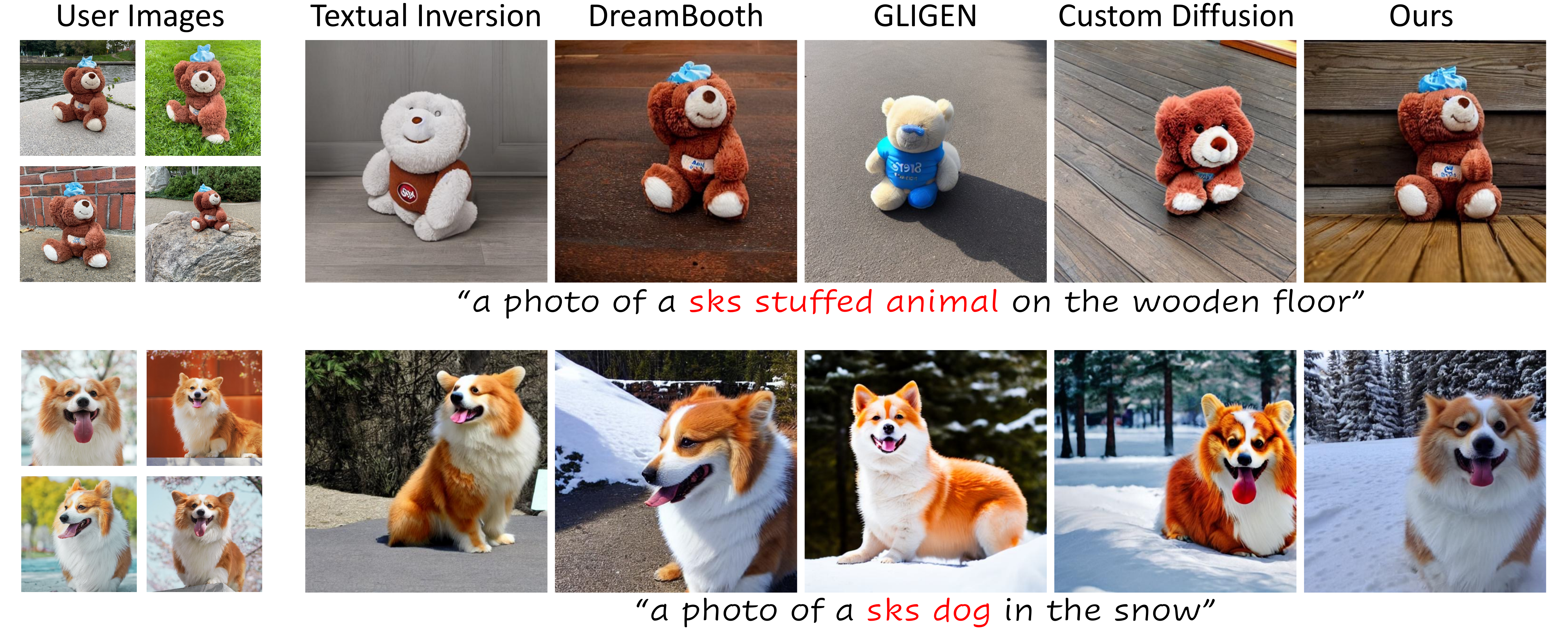}
    \caption{Our method exhibits fidelity comparable to DreamBooth, and outperforms other baselines. }
    \label{fig:compare}
\end{figure*}

Table~\ref{table:main} shows that our model, PACGen, can produce results that are comparable or slightly better in image fidelity than alternative state-of-the-art baselines. The IOU metric demonstrates that our model effectively adheres to the input bounding boxes with the same controllability as GLIGEN, \emph{a capability not present in previous personalized diffusion models}. Figure~\ref{fig:compare} presents a qualitative comparison. It can be observed that our method and DreamBooth are superior in fidelity than the other baselines. In general, we notice that as the number of parameters used for finetuning increases, the fidelity of the learned model improves (Ours $\approx$ Dreambooth > Custom Diffusion > GLIGEN). We hypothesize that the CLIP image encoder, employed for calculating the alignment scores, might not effectively capture fine-grained details that are crucial for assessing personalized objects, which is why the quantitative results do not sufficiently reflect the larger discrepancies revealed in the qualitative results. 


\vspace{-1pt}
\subsection{Multi Object Results}
\vspace{-1pt}

We next evaluate multi-object generation under three configurations: (1) man-made object + man-made object (\texttt{toy cat} + \texttt{clock}); (2) man-made object + living pet (\texttt{backpack\_dog} + \texttt{cat}); and (3) living pet + living pet (\texttt{cat2} + \texttt{dog2}). Following~\cite{multiconcepy} we use 8 prompts, where each prompt is used to generate 50 images for evaluation. Table~\ref{table:main} shows a comparison with Custom Diffusion. Our method attains slightly inferior text alignment, which could be due to our fine-tuning the entire model, while the baseline fine-tunes only a portion of it, resulting in reduced overfitting. Nevertheless, our method demonstrates improved fidelity and offers location control, a capability absent in~\cite{multiconcepy}.



\begin{figure*}[t!]
    \centering
    \includegraphics[width=\textwidth]{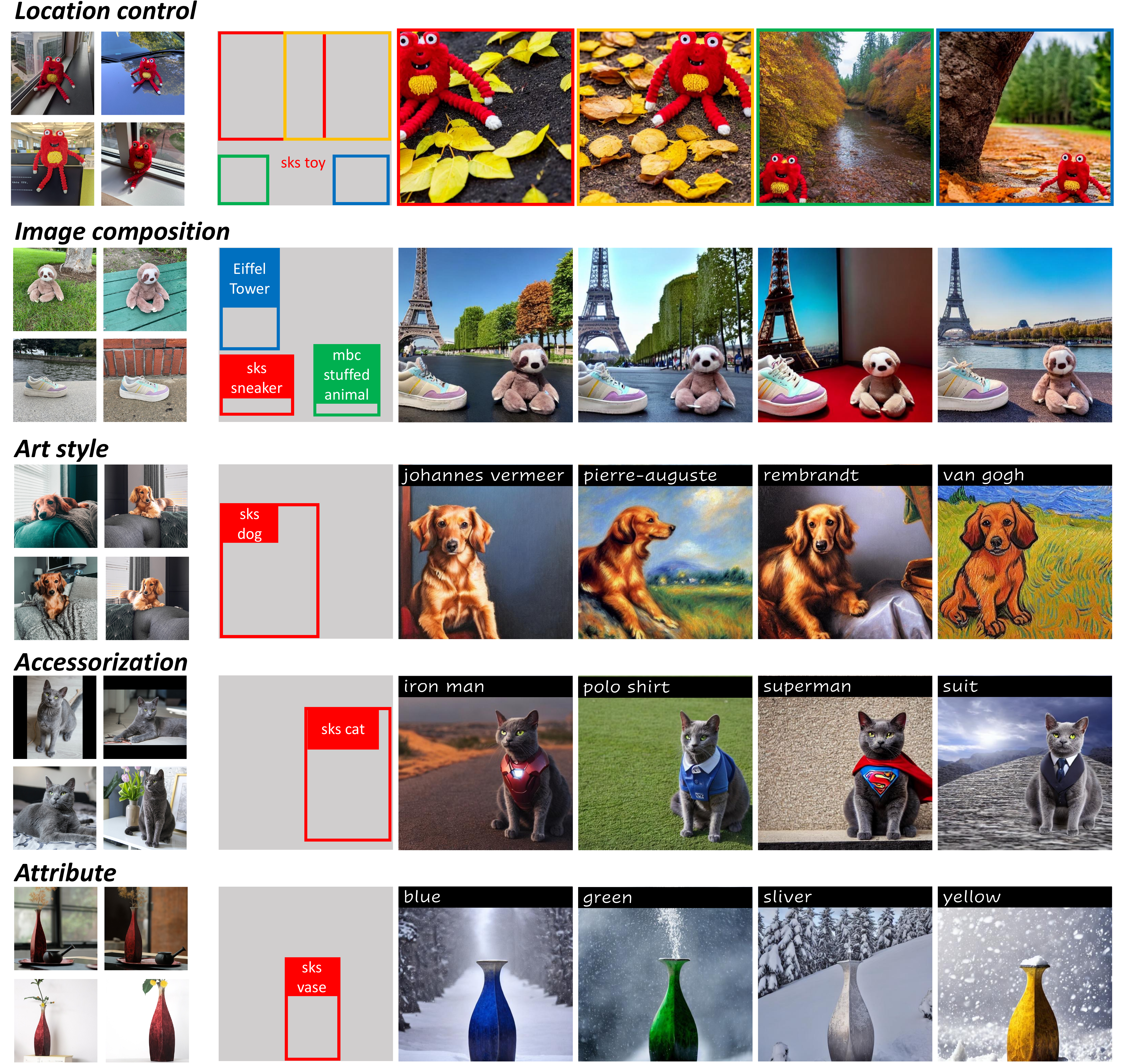}
    \caption{Our method offers versatile applications. It can generate objects at specified locations in diverse scenes and artistic styles, and accessorize objects or modify their attributes with precision. }
    \label{fig:application}
\end{figure*}

\vspace{-1pt}
\subsection{Ablation Study and Applications} 
\vspace{-1pt}

\textbf{Finetuning GLIGEN's controllable layers.}
In the training phase, our method finetunes a pretrained text-to-image diffusion model by utilizing data augmentation. In the inference phase, we plug-and-play new layers from a pre-trained GLIGEN model (refer to Eq.~\ref{eq:gated-self-attention}) without further finetuning. If we also finetune the GLIGEN layers, we obtain 0.786, 0.778, and 0.834 for the text, image, and object alignment scores, respectively. These values are comparable to the numbers in Table~\ref{table:main}. This suggests that the original diffusion model possesses sufficient capacity for visual concept representation, and the new layers introduced by GLIGEN only serve the purpose of spatially adjusting the visual features according to the inserted controllable location and size information (e.g., bounding boxes). 


\textbf{Fidelity to identity at different scales and locations.} The bounding boxes utilized for our evaluation in Table~\ref{table:main} were automatically derived from DreamBooth results, and can be considered as having `typical' object size and location. To investigate the impact of varying the size and position of bounding boxes on the fidelity of the generated object identity, we assess our method by scaling the boxes across scales [0.5, 
 0.6, 0.7, 0.8, 0.9, \textbf{1.0}, 1.2, 1.5, 2.0] and random positions. We evaluate the object alignment score, as text and image alignment are not appropriate measures due to the fact that the object occupies only a small portion of images at smaller scales. The results are [0.78, 0.80, 0.81, 0.82, 0.83, \textbf{0.83}, 0.82, 0.81, 0.78], which demonstrate that our method can maintain object fidelity across a wide range of bounding box scales and locations.
 

\textbf{Regional Guided Sampling.} It is introduced to address artifacts leaked by data augmentation. We ablate the effect of each component. (1) Removing regular negative prompt $p_{\mathrm{negative}}$, which is set as \texttt{``collaging effect, low quality, assembled image''}. We train a binary classifier (see supp for details) to classify if collaging effects exist in an image. The percentage of collaging examples are 12.7\% and 2.4\% before and after adding this prompt. (2) Removing multi-object suppression prompt $p_{\mathrm{suppress}}$. We again use~\cite{SAM} to detect objects and use mIOU for evaluation. The result is dropped from 0.795 to 0.702 due to multiple generated objects (vs only one GT box). (3) Removing diversity prompt $p_{\mathrm{diversity}}$. We calculate LPIPS~\cite{lpips} distance between pairs of images sharing the same prompt. The distance drops from 0.6583 to 0.5489 (i.e., the generated images become less diverse).  We show random qualitative samples for these three ablations in the supp.


\textbf{Applications.} Fig.~\ref{fig:application} showcases various intriguing applications of PACGen. The \textbf{Composition} feature allows for the creation of multiple objects from user images with high fidelity and the generation of additional objects (e.g., Eiffel Tower) with location control. The \textbf{Art Style} aspect enables art style changes for personalized objects while maintaining location control. Lastly, \textbf{Accessorization} and \textbf{Attribute} features facilitate the addition of accessories or the modification of attributes (e.g., color) for personalized objects, allowing them to be freely positioned at any location within a scene.

\vspace{-5pt}
\section{Conclusion and Broader Impact}
\vspace{-5pt}

We introduced a straightforward yet effective approach for addressing the entanglement issue in existing personalization diffusion models, and incorporated a inference-time sampling technique to ensure high-quality generation with location control. Nevertheless, our work has some limitations. We occasionally observed some attribute mixture in multi-concept cases; for example, the color of one object can blend into another object. Our method takes more time for inference due to two additional prompts ($p_{\mathrm{suppress}}$ and $p_{\mathrm{diversity}}$) in the diffusion process, with the inference time increasing from 5 to 8 seconds per image on an A6000 GPU.

This project aims to provide users with a tool for controllable synthesis of personalized subjects in various contexts, offering improved location control. However, the potential misuse of generated images by bad actors remains a concern. Future research should continue to address these issues in generative modeling and personalized generative priors.

\appendix
\section*{Appendix}
\renewcommand{\thesection}{\Alph{section}}

\textbf{Training details.} We implement our training based on the DreamBooth code from the diffusers~\cite{diffusers}. We finetune Stable Diffusion v1.4~\cite{LDM} with batch size as 2 for 1000 iterations. We set random resizing scalar $s$ (Eq. 6) as 0.3 for all experiments. All other hyperparamters are the same as the default ones in diffusers~\cite{diffusers}. For multi concepts, we double the batch size to 4 following~\cite{multiconcepy}.   

\textbf{Inference details.} In the inference phase, we integrate location control adapters (Eq~\ref{eq:gated-self-attention}) from GLIGEN into the finetuned Stable Diffusion model. The classifier-free guidance scale is set to 5 (Eq~\ref{eq:prompt}), and we employ schedule sampling with a value of 0.3 for $\tau$, which is a technique proposed by GLIGEN~\cite{gligen} that helps strike a balance between image quality and layout correspondence.   

\textbf{Collaging classifer.} In order to evaluate the effects of collaging in our ablation study, we train a classifier. Specifically, we utilize the CLIP~\cite{CLIP} ViT-L-14 variant to extract features from images and subsequently train a linear classifier using these features while keeping the backbone fixed. Our dataset consists of ADE20K images. Each regular image is treated as a non-collaged class, while two randomly selected images are combined via collaging (by randomly resizing and positioning one image onto the other) to create a collaged class. We train the linear classification head using a batch size of 128 for 50 epochs.

\section{Additional qualitative results}

\begin{figure*}[h!]
\vspace{-20pt}
    \centering
    \includegraphics[width=0.9\textwidth]{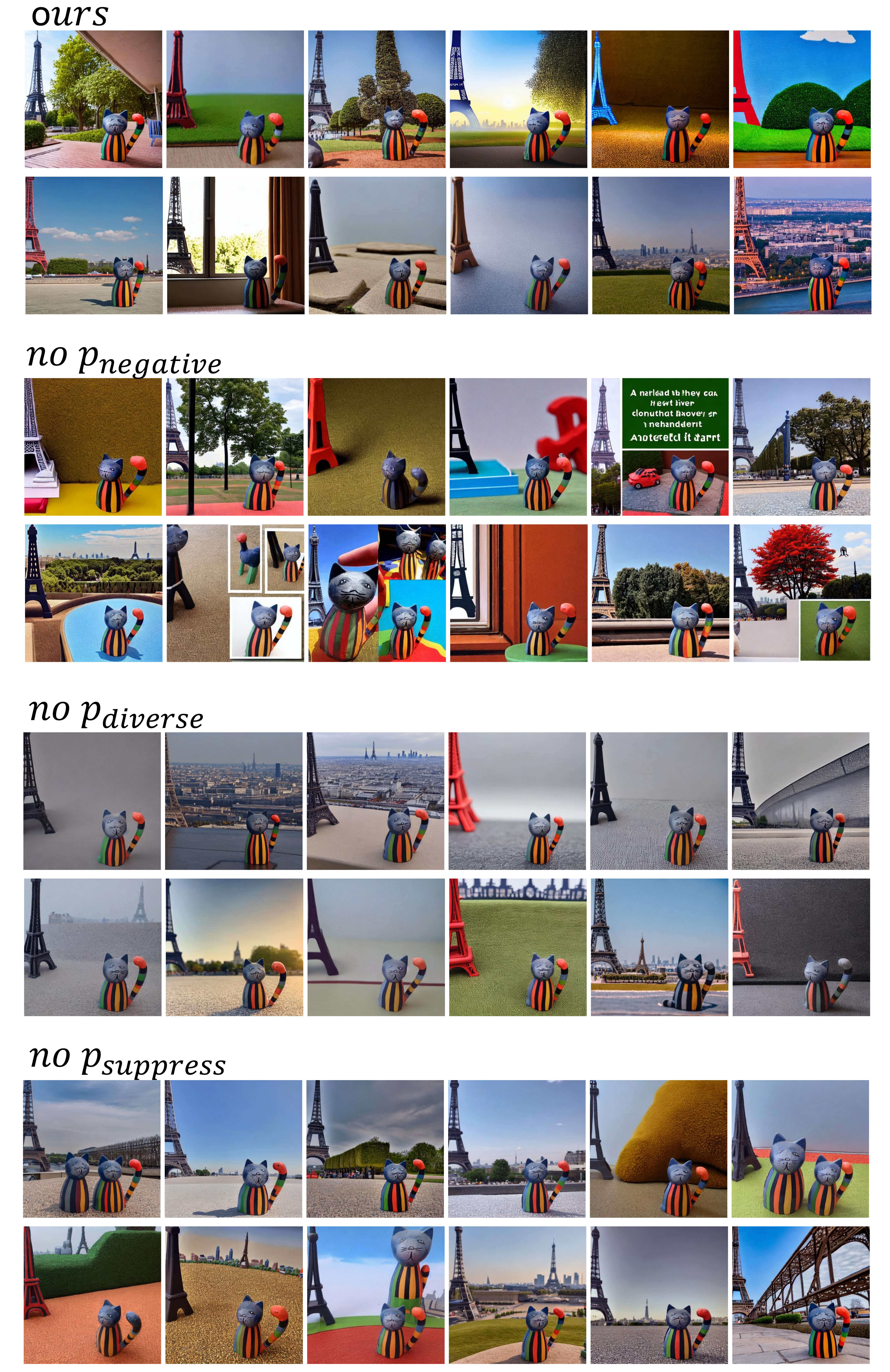}
    \caption{ \textbf{Qualitative ablation.} Random samples. The caption and layout is the same as Fig.~5 in the main paper.}
    \label{fig:ablation}
\end{figure*}

Apart from conducting a quantitative ablation study as in the main paper, we also present a visual ablation study for our designs. The crucial designs during the inference stage involve regional guidance, which comprises three prompts. We systematically ablate each prompt, and Figure~\ref{fig:ablation} displays random samples obtained by removing each of them. The results clearly demonstrate the necessity of each prompt. When $p_{\mathrm{negative}}$ is absent, the results exhibit a collaging effect, especially when the object only occupies a relatively small part of the image, as illustrated in the provided case. Furthermore, the results appear less colorful and less interesting in the absence of $p_{\mathrm{negative}}$. Lastly, without $p_{\mathrm{suppress}}$, the model tends to generate additional instances in undesired locations.

\begin{figure*}[h!]
    \centering
    \includegraphics[width=0.9\textwidth]{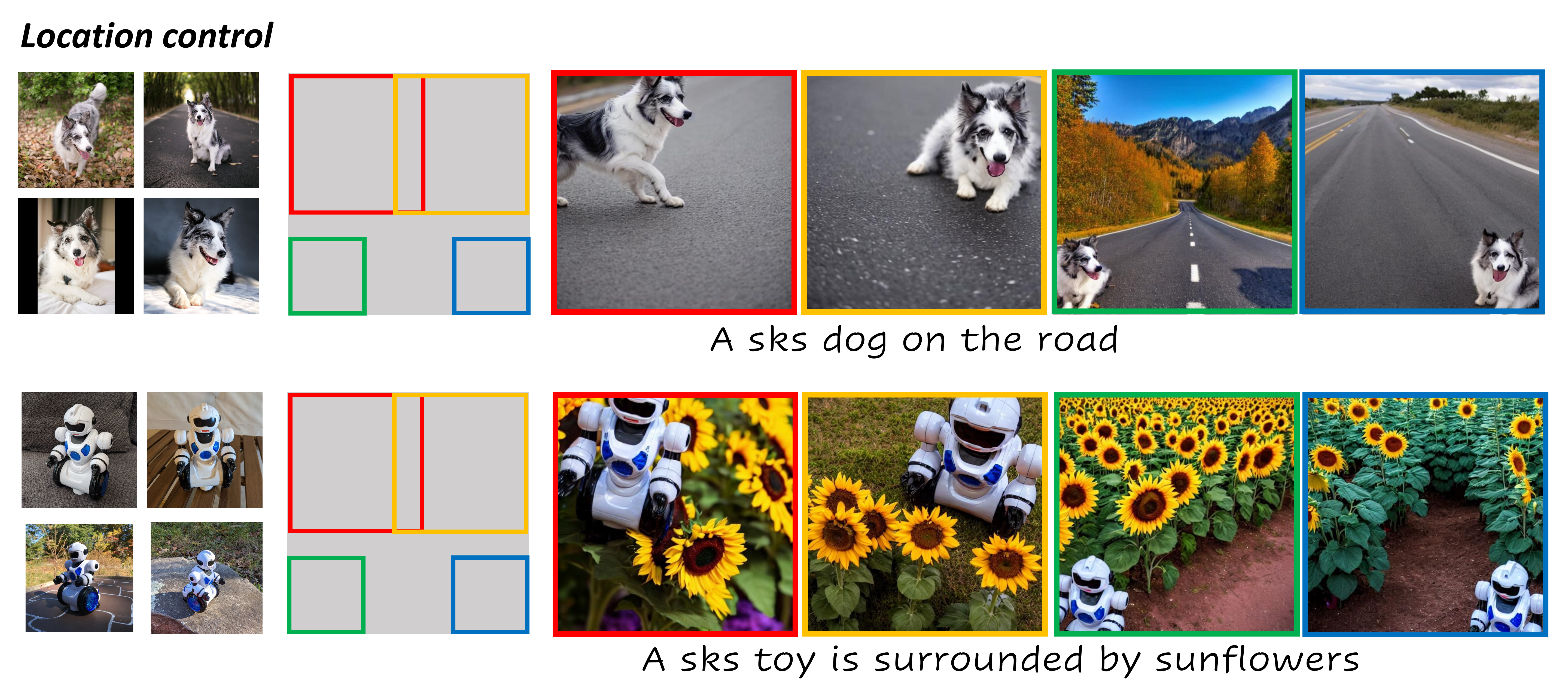}
    \caption{ \textbf{Location control.} For the same concept, our model can position it in any arbitrary location.}
    \label{fig:location}
\end{figure*}

\begin{figure*}[h!]
    \centering
    \includegraphics[width=0.9\textwidth]{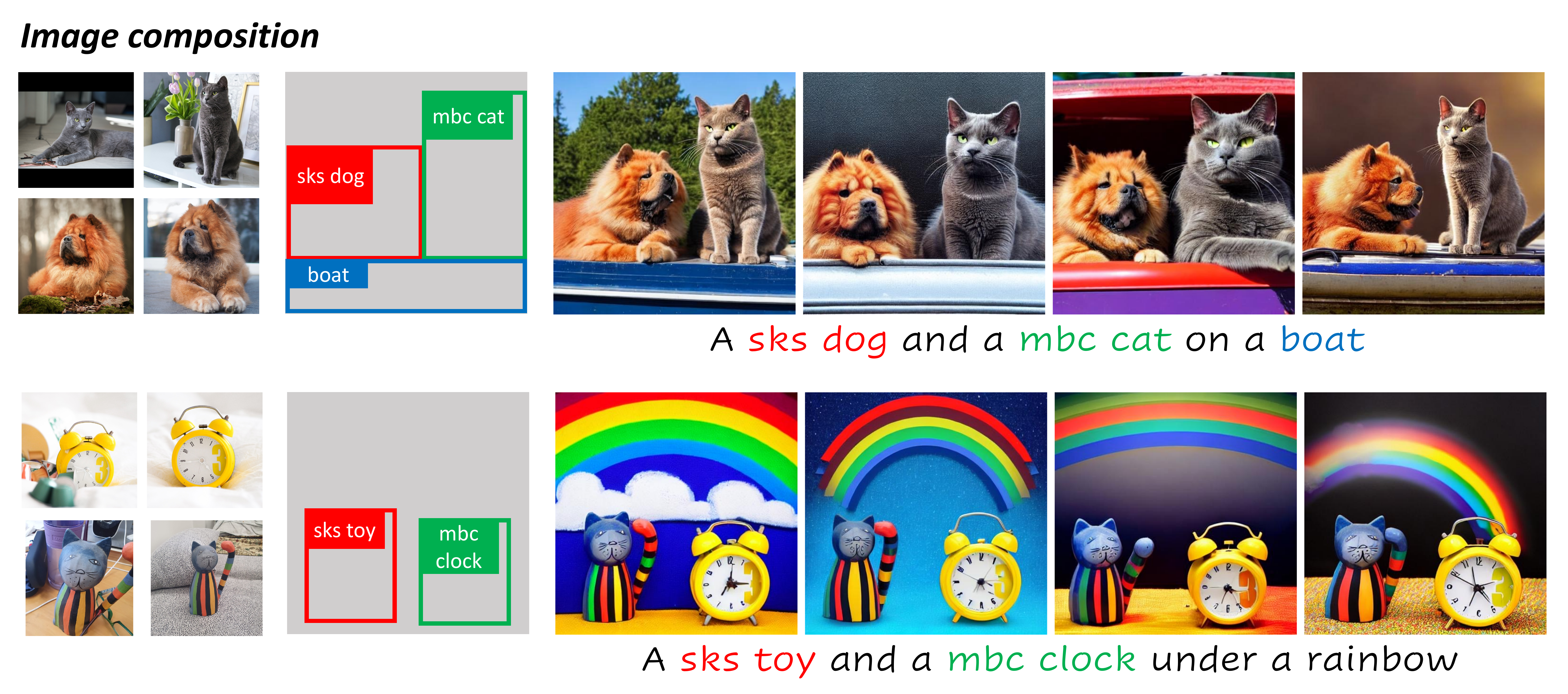}
    \caption{ \textbf{Multi object composition.} Our model can combine multiple instances provided by a user.}
    \label{fig:composition}
\end{figure*}

\begin{figure*}[h!]
    \centering
    \includegraphics[width=0.9\textwidth]{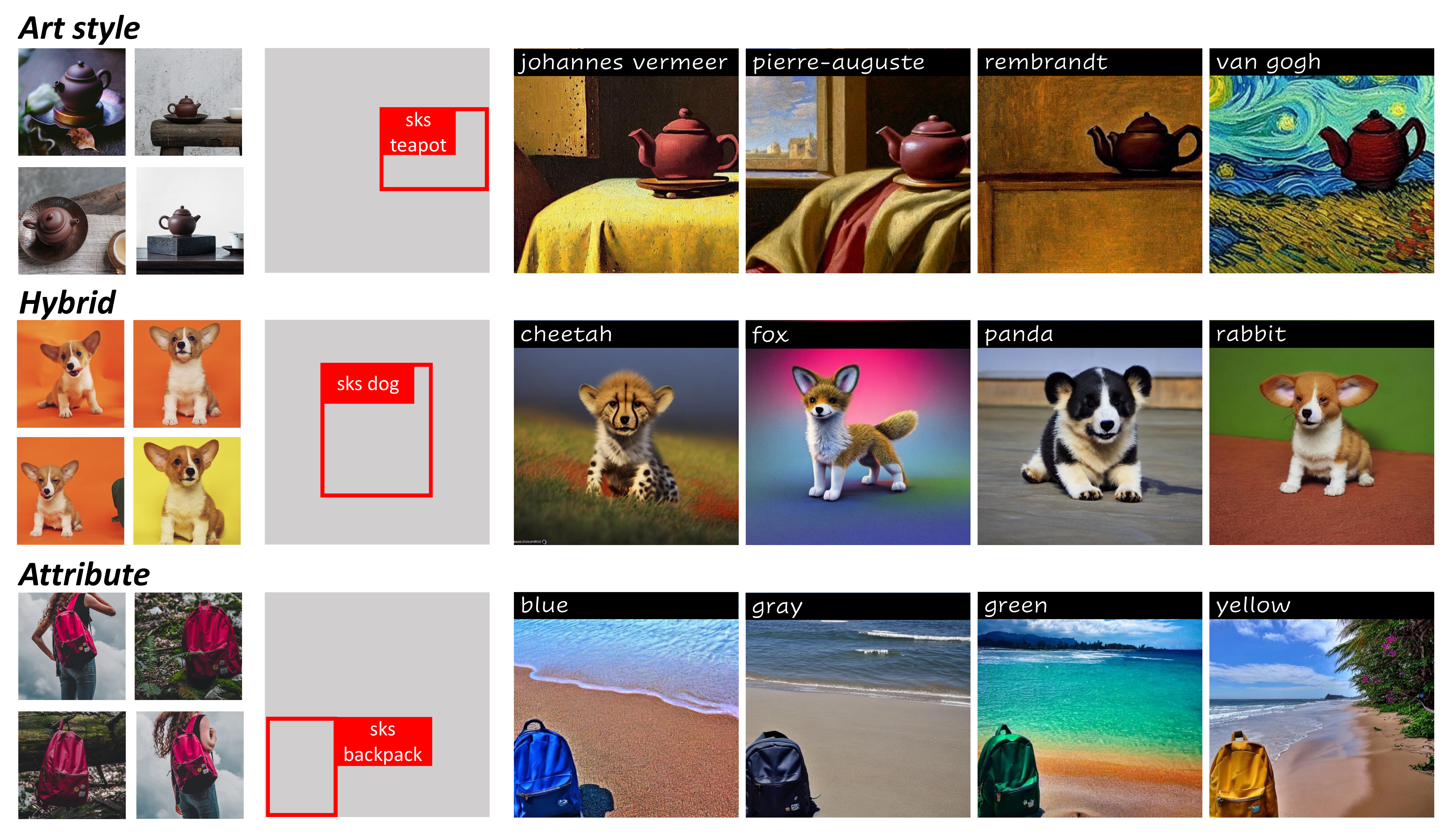}
    \caption{ \textbf{Applications.} We enable various applications while controlling the location and size of objects.}
    \label{fig:art_hybrid_attribute}
\end{figure*}

Finally, Figure~\ref{fig:location}, Figure~\ref{fig:composition}, and Figure~\ref{fig:art_hybrid_attribute} show additional results on single/multi-object location control and various applications by our model PACGen.

\clearpage

{
\bibliographystyle{plain}
\bibliography{ref}

\begin{thebibliography}{10}

\bibitem{universal-bansal-arxiv2023}
Arpit Bansal, Hong-Min Chu, Avi Schwarzschild, Soumyadip Sengupta, Micah
  Goldblum, Jonas Geiping, and Tom Goldstein.
\newblock Universal guidance for diffusion models.
\newblock {\em arXiv preprint arXiv:2302.07121}, 2023.

\bibitem{kid}
Mikolaj Binkowski, Danica~J. Sutherland, Michal Arbel, and Arthur Gretton.
\newblock Demystifying mmd gans.
\newblock {\em arXiv}, abs/1801.01401, 2018.

\bibitem{bigGAN}
Andrew Brock, Jeff Donahue, and Karen Simonyan.
\newblock Large scale gan training for high fidelity natural image synthesis.
\newblock {\em arXiv preprint arXiv:1809.11096}, 2018.

\bibitem{ding2021cogview}
Ming Ding, Zhuoyi Yang, Wenyi Hong, Wendi Zheng, Chang Zhou, Da~Yin, Junyang
  Lin, Xu~Zou, Zhou Shao, Hongxia Yang, and Jie Tang.
\newblock Cogview: Mastering text-to-image generation via transformers.
\newblock {\em NeurIPS}, 2021.

\bibitem{textualinv}
Rinon Gal, Yuval Alaluf, Yuval Atzmon, Or~Patashnik, Amit~H Bermano, Gal
  Chechik, and Daniel Cohen-Or.
\newblock An image is worth one word: Personalizing text-to-image generation
  using textual inversion.
\newblock {\em arXiv preprint arXiv:2208.01618}, 2022.

\bibitem{GANs2016}
Ian Goodfellow, Jean Pouget-Abadie, Mehdi Mirza, Bing Xu, David Warde-Farley,
  Sherjil Ozair, and Yoshua Bengio.
\newblock Generative adversarial networks.
\newblock {\em NeurIPS}, 2014.

\bibitem{cfg}
Jonathan Ho.
\newblock Classifier-free diffusion guidance.
\newblock {\em arXiv}, abs/2207.12598, 2022.

\bibitem{ddpm}
Jonathan Ho, Ajay Jain, and Pieter Abbeel.
\newblock Denoising diffusion probabilistic models.
\newblock {\em NeurIPS}, 2020.

\bibitem{Composer}
Lianghua Huang, Di~Chen, Yu~Liu, Yujun Shen, Deli Zhao, and Jingren Zhou.
\newblock Composer: Creative and controllable image synthesis with composable
  conditions.
\newblock {\em arXiv preprint arXiv:2302.09778}, 2023.

\bibitem{styleGAN}
Tero Karras, Samuli Laine, and Timo Aila.
\newblock A style-based generator architecture for generative adversarial
  networks.
\newblock In {\em CVPR}, 2019.

\bibitem{SAM}
Alexander Kirillov, Eric Mintun, Nikhila Ravi, Hanzi Mao, Chloe Rolland, Laura
  Gustafson, Tete Xiao, Spencer Whitehead, Alexander~C. Berg, Wan-Yen Lo, Piotr
  Doll{\'a}r, and Ross Girshick.
\newblock Segment anything.
\newblock {\em arXiv:2304.02643}, 2023.

\bibitem{multiconcepy}
Nupur Kumari, Bingliang Zhang, Richard Zhang, Eli Shechtman, and Jun-Yan Zhu.
\newblock Multi-concept customization of text-to-image diffusion.
\newblock {\em CVPR}, 2023.

\bibitem{lee2019countering}
Jason Lee, Kyunghyun Cho, and Douwe Kiela.
\newblock Countering language drift via visual grounding.
\newblock In {\em EMNLP}, 2019.

\bibitem{gligen}
Yuheng Li, Haotian Liu, Qingyang Wu, Fangzhou Mu, Jianwei Yang, Jianfeng Gao,
  Chunyuan Li, and Yong~Jae Lee.
\newblock Gligen: Open-set grounded text-to-image generation.
\newblock {\em CVPR}, 2023.

\bibitem{coco}
Tsung-Yi Lin, Michael Maire, Serge~J. Belongie, James Hays, Pietro Perona, Deva
  Ramanan, Piotr Doll{\'a}r, and C.~Lawrence Zitnick.
\newblock Microsoft coco: Common objects in context.
\newblock In {\em ECCV}, 2014.

\bibitem{lu2020countering}
Yuchen Lu, Soumye Singhal, Florian Strub, Aaron Courville, and Olivier
  Pietquin.
\newblock Countering language drift with seeded iterated learning.
\newblock In {\em ICML}, 2020.

\bibitem{t2i}
Chong Mou, Xintao Wang, Liangbin Xie, Jian Zhang, Zhongang Qi, Ying Shan, and
  Xiaohu Qie.
\newblock T2i-adapter: Learning adapters to dig out more controllable ability
  for text-to-image diffusion models.
\newblock {\em arXiv preprint arXiv:2302.08453}, 2023.

\bibitem{GLIDE}
Alex Nichol, Prafulla Dhariwal, Aditya Ramesh, Pranav Shyam, Pamela Mishkin,
  Bob McGrew, Ilya Sutskever, and Mark Chen.
\newblock Glide: Towards photorealistic image generation and editing with
  text-guided diffusion models.
\newblock {\em ICML}, 2022.

\bibitem{CLIP}
Alec Radford, Jong~Wook Kim, Chris Hallacy, Aditya Ramesh, Gabriel Goh,
  Sandhini Agarwal, Girish Sastry, Amanda Askell, Pamela Mishkin, Jack Clark,
  Gretchen Krueger, and Ilya Sutskever.
\newblock Learning transferable visual models from natural language
  supervision.
\newblock In {\em ICML}, 2021.

\bibitem{T5}
Colin Raffel, Noam Shazeer, Adam Roberts, Katherine Lee, Sharan Narang, Michael
  Matena, Yanqi Zhou, Wei Li, and Peter~J Liu.
\newblock Exploring the limits of transfer learning with a unified text-to-text
  transformer.
\newblock {\em Journal of Machine Learning Research}, 21(1):5485--5551, 2020.

\bibitem{DALLE2}
Aditya Ramesh, Prafulla Dhariwal, Alex Nichol, Casey Chu, and Mark Chen.
\newblock Hierarchical text-conditional image generation with clip latents.
\newblock {\em arXiv preprint arXiv:2204.06125}, 2022.

\bibitem{ramesh2021zero}
Aditya Ramesh, Mikhail Pavlov, Gabriel Goh, Scott Gray, Chelsea Voss, Alec
  Radford, Mark Chen, and Ilya Sutskever.
\newblock Zero-shot text-to-image generation.
\newblock In {\em ICML}, 2021.

\bibitem{LDM}
Robin Rombach, A.~Blattmann, Dominik Lorenz, Patrick Esser, and Bj{\"o}rn
  Ommer.
\newblock High-resolution image synthesis with latent diffusion models.
\newblock {\em CVPR}, 2022.

\bibitem{ruiz2022dreambooth}
Nataniel Ruiz, Yuanzhen Li, Varun Jampani, Yael Pritch, Michael Rubinstein, and
  Kfir Aberman.
\newblock Dreambooth: Fine tuning text-to-image diffusion models for
  subject-driven generation.
\newblock {\em arXiv preprint arxiv:2208.12242}, 2022.

\bibitem{Imagen}
Chitwan Saharia, William Chan, Saurabh Saxena, Lala Li, Jay Whang, Emily~L
  Denton, Kamyar Ghasemipour, Raphael Gontijo~Lopes, Burcu Karagol~Ayan, Tim
  Salimans, Jonathan Ho, David Fleet, and Mohammad Norouzi.
\newblock Photorealistic text-to-image diffusion models with deep language
  understanding.
\newblock {\em NeurIPS}, 2022.

\bibitem{laion}
Christoph Schuhmann, Richard Vencu, Romain Beaumont, Robert Kaczmarczyk,
  Clayton Mullis, Aarush Katta, Theo Coombes, Jenia Jitsev, and Aran
  Komatsuzaki.
\newblock Laion-400m: Open dataset of clip-filtered 400 million image-text
  pairs.
\newblock {\em ArXiv}, abs/2111.02114, 2021.

\bibitem{DM}
Jascha~Narain Sohl-Dickstein, Eric~A. Weiss, Niru Maheswaranathan, and Surya
  Ganguli.
\newblock Deep unsupervised learning using nonequilibrium thermodynamics.
\newblock In {\em ICML}, 2015.

\bibitem{ddim}
Jiaming Song, Chenlin Meng, and Stefano Ermon.
\newblock Denoising diffusion implicit models.
\newblock {\em arXiv preprint arXiv:2010.02502}, 2020.

\bibitem{vaswani2017attention}
Ashish Vaswani, Noam Shazeer, Niki Parmar, Jakob Uszkoreit, Llion Jones,
  Aidan~N Gomez, {\L}ukasz Kaiser, and Illia Polosukhin.
\newblock Attention is all you need.
\newblock {\em NeurIPS}, 2017.

\bibitem{diffusers}
Patrick von Platen, Suraj Patil, Anton Lozhkov, Pedro Cuenca, Nathan Lambert,
  Kashif Rasul, Mishig Davaadorj, and Thomas Wolf.
\newblock Diffusers: State-of-the-art diffusion models.
\newblock \url{https://github.com/huggingface/diffusers}, 2022.

\bibitem{ganinv}
Weihao Xia, Yulun Zhang, Yujiu Yang, Jing-Hao Xue, Bolei Zhou, and Ming-Hsuan
  Yang.
\newblock Gan inversion: A survey.
\newblock {\em TPAMI}, 2022.

\bibitem{xu2018attngan}
Tao Xu, Pengchuan Zhang, Qiuyuan Huang, Han Zhang, Zhe Gan, Xiaolei Huang, and
  Xiaodong He.
\newblock Attngan: Fine-grained text to image generation with attentional
  generative adversarial networks.
\newblock In {\em CVPR}, 2018.

\bibitem{reco}
Zhengyuan Yang, Jianfeng Wang, Zhe Gan, Linjie Li, Kevin Lin, Chenfei Wu, Nan
  Duan, Zicheng Liu, Ce~Liu, Michael Zeng, and Lijuan Wang.
\newblock Reco: Region-controlled text-to-image generation.
\newblock In {\em CVPR}, 2023.

\bibitem{VQGAN}
Jiahui Yu, Xin Li, Jing~Yu Koh, Han Zhang, Ruoming Pang, James Qin, Alexander
  Ku, Yuanzhong Xu, Jason Baldridge, and Yonghui Wu.
\newblock Vector-quantized image modeling with improved vqgan.
\newblock {\em arXiv preprint arXiv:2110.04627}, 2021.

\bibitem{stackgan}
Han Zhang, Tao Xu, Hongsheng Li, Shaoting Zhang, Xiaogang Wang, Xiaolei Huang,
  and Dimitris~N Metaxas.
\newblock Stackgan: Text to photo-realistic image synthesis with stacked
  generative adversarial networks.
\newblock In {\em ICCV}, 2017.

\bibitem{stackgan++}
Han Zhang, Tao Xu, Hongsheng Li, Shaoting Zhang, Xiaogang Wang, Xiaolei Huang,
  and Dimitris~N Metaxas.
\newblock Stackgan++: Realistic image synthesis with stacked generative
  adversarial networks.
\newblock {\em TPAMI}, 41(8):1947--1962, 2018.

\bibitem{controlnet}
Lvmin Zhang and Maneesh Agrawala.
\newblock Adding conditional control to text-to-image diffusion models.
\newblock {\em arXiv preprint arXiv:2302.05543}, 2023.

\bibitem{lpips}
Richard Zhang, Phillip Isola, Alexei~A Efros, Eli Shechtman, and Oliver Wang.
\newblock The unreasonable effectiveness of deep features as a perceptual
  metric.
\newblock In {\em CVPR}, 2018.

\bibitem{zhu2019dm}
Minfeng Zhu, Pingbo Pan, Wei Chen, and Yi~Yang.
\newblock Dm-gan: Dynamic memory generative adversarial networks for
  text-to-image synthesis.
\newblock In {\em CVPR}, 2019.

\end{thebibliography}
}

\end{document}